%% file: main.tex
\documentclass[twoside,11pt]{article}

\usepackage{jmlr2e}

\usepackage{amsmath}
\usepackage{booktabs}
\usepackage{algorithm}
\usepackage{algorithmic}
\usepackage{makecell}
\usepackage{color}
\usepackage{graphicx} 
\usepackage{subfig}
\usepackage{caption}
\usepackage{float}

\usepackage{amsmath}

\usepackage{graphics}
\usepackage{subfig}
\usepackage{float}
\usepackage{adjustbox}
\usepackage{nicefrac}       % compact symbols for 1/2, etc.
\usepackage{microtype}      % microtypography
\usepackage{amssymb }
\renewcommand{\vec}[1]{\mathbf{#1}}
\usepackage{algorithm}
\usepackage{algorithmic}
\usepackage{color}

\begin{document}

\title{Scalable Combinatorial Bayesian Optimization with Tractable Statistical models}
\author{\name  Aryan Deshwal \email aryan.deshwal@wsu.edu \\
       \addr Washington State University
        \AND
    \name Syrine Belakaria \email syrine.belakaria@wsu.edu \\
       \addr Washington State University
       \AND
       \name Janardhan Rao Doppa \email jana.doppa@wsu.edu\\
       \addr Washington State University
}
       
\maketitle

\begin{abstract}
We study the problem of optimizing expensive blackbox functions over combinatorial spaces (e.g., sets, sequences, trees, and graphs). BOCS \cite{bocs} is a state-of-the-art Bayesian optimization method for tractable statistical models, which performs semi-definite programming based acquisition function optimization (AFO) to select the next structure for evaluation. Unfortunately, BOCS scales poorly for large number of binary and/or categorical variables. Based on recent advances in submodular relaxation \cite{shinji_neurips} for solving Binary Quadratic Programs, we study an approach referred as {\em Parametrized Submodular Relaxation (PSR)} towards the goal of improving the scalability and accuracy of solving AFO problems for BOCS model. PSR approach relies on two key ideas. First, reformulation of AFO problem as submodular relaxation with some unknown parameters, which can be solved efficiently using minimum graph cut algorithms. Second, construction of an optimization problem to estimate the unknown parameters with close approximation to the true objective. Experiments on diverse benchmark problems show significant improvements with PSR for BOCS model. The source code is available at \url{https://github.com/aryandeshwal/Submodular_Relaxation_BOCS}. 
\end{abstract}

\input{files/introduction.tex}
\input{files/related_work.tex}
\input{files/technical_section.tex}

\input{files/experiments.tex}

\section{Conclusions}

This paper studied a principled approach referred as parametrized submodular relaxation (PSR) to improve the scalability and accuracy of the state-of-the-art combinatorial Bayesian optimization algorithm with tractable statistical models called BOCS. The key idea is to reformulate the acquisition function optimization to select the next structure for evaluation as submodular relaxation with some parameters, and perform search over these parameters to improve the accuracy of this relaxed problem. Our experimental results on diverse benchmarks showed that PSR algorithm significantly improved the computational-efficiency and accuracy of BOCS.

\vspace{2.0ex}

\noindent {\bf Acknowledgements.} The first author would like to thank Shinji Ito for useful discussions related to this research. The authors gratefully acknowledge the support from National Science Foundation (NSF) grants IIS-1845922 and OAC-1910213. The views expressed are those of the authors and do not reflect the official policy or position of the NSF.

\bibliography{references}

\end{document}

%% file: files/introduction.tex
\section{Introduction}

Many real-world science and engineering applications involve optimizing combinatorial spaces (e.g., sets, sequences, trees, and graphs) using expensive black-box evaluations. For example, molecular optimization guided by physical lab experiments, where molecules are represented as graphs or SMILE strings \cite{reduction_continuous}. Similarly, in hardware design optimization, we need to appropriately place the processing elements and communication links for achieving high performance guided by expensive computational simulations to emulate the real hardware. 

Bayesian optimization (BO) \cite{BO-Survey:2016} is a popular framework for solving expensive black-box optimization problem. BO framework consists of three key elements: {\em 1) Statistical model} (e.g., Gaussian Process) learned from past function evaluations; {\em 2) Acquisition function (AF)} (e.g., expected improvement) to score the potential utility of evaluating an input based on the statistical model; and {\em 3) Acquisition function optimization (AFO)} to select the best candidate input for evaluation. In each BO iteration, the selected input is evaluated and the statistical model is updated using the aggregate training data. Most of the prior work on BO is focused on optimizing continuous spaces. There are two key challenges to extend BO framework to combinatorial spaces. First, defining a surrogate statistical model over combinatorial objects. Second, search through the combinatorial space to select the next structure for evaluation given such a statistical model. 

Prior work on combinatorial BO has addressed these two challenges as follows. SMAC \cite{SMAC,SMAC:TR2010} is a canonical baseline that employs complex statistical model in the form of random forest and executes a {\em hand-designed} local search procedure for optimizing the acquisition function. A recent work referred as COMBO \cite{combo} proposed a novel combinatorial graph representation for discrete spaces which allows using Gaussian process with diffusion kernels. Reduction to continuous BO \cite{reduction_continuous} employs an encoder-decoder architecture to learn continuous representation of combinatorial objects from data and performs BO in this latent space. Unfortunately, this approach requires a {\em large} dataset of combinatorial objects, for learning the latent space, which is impossible to acquire for many real-world applications. BOCS \cite{bocs} is another method that employs a tractable statistical model defined over binary variables and Thompson sampling as the acquisition function. These two choices within BOCS leads to a semi-definite programming (SDP) based solution for solving AFO problems. Unfortunately, BOCS approach scales poorly for large number of binary variables and for categorical variables due to one-hot encoding representation.

Our work is inspired by the success of submodular relaxation based inference methods in the structured prediction literature \cite{bqp_graph_cuts,submodular_cv_1,submodular_cv_2,submodular_cv_3}. In this paper, we employ the submodular relaxation based Binary Quadratic optimization approach proposed in \cite{shinji_neurips} to improve the computational-efficiency and accuracy of solving AFO problems for BOCS model. We refer to this approach as {\em Parametrized Submodular Relaxation (PSR)} algorithm. First, we reformulate the AFO problem as submodular relaxation with some parameters. This relaxed problem can be solved efficiently using minimum graph cut algorithms \cite{bqp_graph_cuts,boykov_kolmogorov_algorithm}. The accuracy of this relaxed problem critically depends on the unknown parameters. Therefore, we solve an outer optimization problem to find the values of unknown parameters with close approximation to the true objective. {\em To the best of our knowledge, this is the first application of submodular relaxation to solve combinatorial BO problems}. Experimental results on diverse real-world benchmarks show the efficacy of PSR to improve the state-of-the-art on combinatorial BO with tractable statistical models in terms of both computational-efficiency and accuracy.

\vspace{1.0ex}

\noindent {\bf Contributions.} The main contributions of this paper are:

\begin{itemize}
    \item By leveraging the recent advances in submodular relaxation, we study the parametrized submodular relaxation approach to improve the scalability and accuracy of solving AFO problems for BOCS, the state-of-the-method for tractable statistical models.
    \item We perform comprehensive experiments on real-world benchmarks to show computational-efficiency and accuracy improvements over existing BOCS method. The source code is available on the GitHub repository \url{https://github.com/aryandeshwal/Submodular_Relaxation_BOCS}.
\end{itemize}

%% file: files/related_work.tex
\section{Problem Setup}

We are given a combinatorial space of structures $\mathcal{X}$ (e.g., sets, sequences, trees, graphs). Without loss of generality, let each combinatorial structure $\vec{x} \in \mathcal{X}$ be represented using $n$ discrete variables $x_1,x_2,\cdots,x_n$, where each variable $x_i$ takes $k$ candidate values from the set $\mathcal{CV}(x_i)$. For binary variables, $k$ equals 2 and for categorical variables, $k$ is greater than 2. We assume the availability of an unknown objective function $\mathcal{F}: \mathcal{X} \mapsto \Re$ to evaluate each combinatorial object $\vec{x} \in \mathcal{X}$. Each evaluation is expensive and results in outcome $y$ = $\mathcal{F}(\vec{x})$. For example, in hardware design optimization, $\vec{x}$ is a graph corresponding to the placement of processing elements and communication links, and $\mathcal{F}(\vec{x})$ corresponds to an expensive computational simulation. The overall goal is to minimize the number of objective function evaluations to uncover a structure $\vec{x} \in \mathcal{X}$ that approximately optimizes $\mathcal{F}$. We consider minimizing the objective $\mathcal{F}$ for the sake of technical exposition and consistent notation.

%\vspace{1.0ex}

\section{Related Work}
There is very limited work on BO over discrete spaces when compared to
continuous space BO, which has seen huge growth over the last few years \cite{bo_tutorial,bo_materialdesign,BO:NIPS2012,MES,MF-IS,hernandez2016predictive,MESMO,USEMO}. SMAC \cite{SMAC,SMAC:TR2010} is one canonical baseline which employs random forest as surrogate model and a {\em hand-designed} local search procedure for optimizing the acquisition function. BOCS \cite{bocs} employs a parametric statistical models over binary variables, which allows pricipled acquisition function optimization based on semi-definite program solvers. COMBO \cite{combo} is a state-of-the-art non-parametric approach that employs Gaussian processes  with diffusion kernels defined over discrete spaces as its surrogate model. However, COMBO employs local search with random restarts for acquisition function optimization. 

COMBO was shown to achieve better performance than BOCS for complex domains that require modeling higher-order dependencies between discrete variables. However, BOCS achieves good performance whenever the modeling assumptions (e.g., lower-order interactions among variables) are met. Furthermore, the AFO problem in BOCS is a Binary Quadratic Programming (BQP) problem which is well-studied in many fields including computer vision \cite{bqp_graph_cuts} and prescriptive price optimization \cite{shinji_neurips}. In comparison, the acquisition function optimization is much more challenging (results in general non-linear combinatorial optimization problem) for methods such as COMBO and SMAC that employs non-parametric statistical models. A learning to search framework referred as L2S-DISCO \cite{l2sdisco} was introduced recently to solve the challenges of AFO problems with complex statistical models (e.g., GPs with discrete kernels and random forest). The key insight behind L2S-DISCO is to directly tune the search via learning during the optimization process to select the next structure for evaluation by leveraging the close relationship between AFO problems across BO iterations (i.e., amortized AFO). Since the main focus of this paper is on improving the scalability and accuracy of AFO for the tractable statistical model introduced in BOCS, we describe the details of this approach below.

\vspace{1.0ex}

\noindent {\bf BOCS Approach.} We now briefly explain the BOCS method \cite{bocs} that we intend to improve on. BOCS instantiates the three key elements of BO framework as follows. {\em 1) Surrogate statistical model:} A linear Bayesian model defined over binary variables is employed as the surrogate model. The model is described as:
\begin{align}
    f_\alpha (\vec{x} \in \mathcal{X}) = \alpha_0 + \sum_j \alpha_j x_j + \sum_{i, j > i} \alpha_{ij} x_i x_j
\end{align}
where $\mathcal{X} = \{0, 1\}^n$ and $\vec{x} \in \mathcal{X}$ is a binary vector and $\alpha$ variables are drawn from a sparsity-inducing horseshoe prior \cite{horseshoe}. It was experimentally found that the above second-order model provides an excellent trade-off between expressiveness and accuracy. The $\alpha$ variables quantify the uncertainty  of the model.  {\em  2) Acquisition function:} Thompson sampling \cite{thompson_sampling} is employed as the acquisition function because of its proven theoretical and empirical properties in the context of BO. 
{\em 3) Acquisition function optimization:} In each BO iteration, we select a candidate structure $\vec{x} \in \mathcal{X}$ for evaluation that minimizes the acquisition function. In the case of BOCS method, the acquisition function optimization (AFO) problem becomes:
\begin{align}
    \arg \min_{\vec{x} \in \mathcal{X}}  f_{\vec{\alpha}}(\vec{x}) + \lambda P(\vec{x}) \label{eqn:afo_bocs_original}%\\
\end{align}
where $\lambda P(\vec{x})$ being a regularization term commonly seen in multiple applications. BOCS employs a semi-definite programming (SDP) based relaxation approach to solve the above AFO problem.

\vspace{1.0ex}

\noindent {\bf Scalability Challenges of BOCS.} There are multiple challenges associated with SDP approach used for solving AFO problems in BOCS formulation. First, the time complexity of a standard SDP solver grows at the rate of $O(n^6)$ \cite{semidefinite,shinji_neurips}, which is prohibitive for large dimensions. Second, the approximation error for SDP based solution is known to be at most $O(\log n)$ \cite{bocs,sdp_accuracy}, which clearly grows as the dimensions increase, resulting in the loss of accuracy as well. These scaling issues arise when the number of binary variables are large. Since BOCS represents categorical variables using one-hot encoding, even a small number of categorical variables can lead to a large number of binary variables (e.g., placement of processing elements in hardware design). 

Our goal in this paper is to provide an algorithmic approach to improve the computational-efficiency and accuracy of solving AFO problems for BOCS method.

%% file: files/technical_section.tex
\section{Parametrized Submodular Relaxation}

In this section, we describe our proposed parameterized submodular relaxation (PSR) algorithm to improve the efficiency and accuracy of solving acquisition function optimization (AFO) problems for BOCS model. We first provide a high-level overview of PSR algorithm. Subsequently, we describe the details of key algorithmic steps behind PSR.

\subsection{High-level Overview of PSR Algorithm}
The overall idea of using submodular relaxations for optimizing BQP problems is extensively employed in computer vision \cite{submodularization_cv}. However, we employ the concrete instantiation of this general framework as proposed in the context of prescriptive price optimization \cite{shinji_neurips}.  {\em To the best of our knowledge, this is the first application of submodular relaxation concepts to solve combinatorial BO problems.}

\vspace{1.0ex}

\noindent Recall that AFO problem for BOCS method is:
\begin{align}
    \arg \min_{\vec{x} \in \mathcal{X}}  f_\vec{\alpha}(\vec{x}) + \lambda P(\vec{x}) \label{eqn:afo}
\end{align}
where $\lambda P(\vec{x})$ is a regularization term commonly seen in multiple applications. For example, by choosing $P(\vec{x}) = \|\vec{x}\|_1$, the optimization problem in \ref{eqn:afo} becomes a Binary Quadratic Program (BQP) as given below:
\begin{align}
    \arg \min_{\vec{x} \in \mathcal{X}} \quad & \alpha_0 + \sum_j (\alpha_j + \lambda) x_j + \sum_{i, j > i} \alpha_{ij}x_i x_j  \\
     \arg \min_{\vec{x} \in \mathcal{X}} \quad & \vec{x}^T A \vec{x} + b^T \vec{x} \label{eqn:main}
\end{align}
In the general case, BQP is NP-hard \cite{sdp_accuracy,nphard_bqp}. We propose using an efficient submodular relaxation with some unknown parameters (matrix $\Lambda$) for solving this problem. A key advantage of this relaxation is that it allows us to leverage minimum graph cut algorithms to efficiently solve it. The accuracy of this relaxed problem critically depends on the unknown parameters $\Lambda$. Therefore, we can utilize an outer optimization problem to find the values of unknown parameters with close approximation to the true acquisition function. We solve this optimization problem using an iterative algorithm (steps 5-9 in Algorithm \ref{alg:PSR}). We perform two algorithmic steps in each iteration. First, we solve the parametrized submodular relaxation of the AFO problem using a minimum graph cut algorithm (step 7). Second, we update the values of unknown parameters $\Lambda$ using proximal gradient descent (step 8). Algorithm \ref{alg:PSR} provides the complete pseudo-code of combinatorial BO using PSR algorithm. 

\vspace{1.0ex}

\noindent {\bf Advantages of PSR algorithm.} When compared to the semi-definite programming (SDP) relaxation approach to solve AFO problems in BOCS, PSR algorithm has significant advantages in terms of both computational-efficiency and accuracy of solving AFO problems. First, PSR relies on a small number of calls (five to ten iterations based on our experiments) to a minimum graph cut solver, which has relatively very low time-complexity (e.g. $\mathcal{O}(n^3)$ for preflow-push algorithms or $\mathcal{O}(n^3 \log n)$ for Dinic's algorithm \cite{network_flows}) which is significantly better than $\mathcal{O}(n^6)$ for SDP approach. Second, our experiments show that PSR algorithm significantly improves the accuracy over SDP approach with increased dimensionality.

\subsection{Key Algorithmic Steps}

The two main algorithmic steps of PSR algorithm are: submodularation of the objective with unknown parameters and finding optimized parameters to improve the accuracy of relaxation. We describe their details below.

\subsubsection{Submodularization of the Objective}

The binary quadratic objective function ($\vec{x}^T A \vec{x} + b^T \vec{x}$ in \ref{eqn:main}) is called as {\em submodular} if all the elements of the matrix $A$ are non-positive, i.e., $a_{ij} \in A \leq 0 \quad \forall i,j$. Such functions are also called as {\em regular} functions in the computer vision community \cite{bqp_graph_cuts}. Note that this definition can also be derived from the popular definition of submodularity defined for set functions. This can be achieved by defining inclusion of an element in a set by $1$ and exclusion by $0$ thereby resulting in an equivalence between binary valued functions and set functions. However, in our setting, it is not necessary that the BQP objective in Equation \ref{eqn:main} follows the submodularity property. As mentioned earlier, we approximate the objective by constructing a submodular relaxation in the same way as \cite{shinji_neurips}. This relaxation is parametrized by a matrix $\Lambda$, which is directly related to the quality of approximation.

\begin{algorithm}[t]
\caption{Combinatorial BO via PSR Algorithm}
\footnotesize
\textbf{Input}: $\mathcal{X}$ = Discrete space, 
$\mathcal{F}(\vec{x})$ = expensive objective function, statistical model $f_{\alpha}(\vec{x} \in \mathcal{X})$ \\
\textbf{Output}: ($\vec{x}_{best}$, $\mathcal{F}(\vec{x}_{best})$), the best uncovered input $x_{best}$ with its corresponding function value
\label{alg:PSR}
\begin{algorithmic}[1]
\STATE Initialize statistical model $f_{\alpha}$ with a  small number of input-output examples; and $t \leftarrow$ 0
\REPEAT
\STATE  Sample $\alpha$ from posterior of $f_{\alpha}$ 
\STATE \textcolor{red}{ Compute the next input to evaluate via acquisition function optimization: \\ $\vec{x}_{t+1} \leftarrow \mbox{arg}\min_{\vec{x} \in \mathcal{X}} \, {AF}(f_{\alpha},\vec{x})$}
\begin{ALC@g}
\STATE  \textcolor{red}{Initialize parameters $\Lambda$}
\REPEAT 
\STATE \textcolor{red}{Solve parametrized submodular relaxation of the AFO problem using graph cuts}
\STATE \textcolor{red}{Update $\Lambda$ via proximal gradient descent}
\UNTIL{\textcolor{red} {convergence of optimization over $\Lambda$}}
\end{ALC@g}
\STATE Evaluate objective function $\mathcal{F}(\vec{x})$ at $\vec{x}_{t+1}$ to get $y_{t+1}$
\STATE Aggregate the data: $\mathcal{D}_{t+1} \leftarrow \mathcal{D}_{t} \cup \left\{(x_{t+1}, y_{t+1})\right\}$ and update the model
\STATE $t \leftarrow t+1$
\UNTIL{convergence or maximum iterations}
\STATE $\vec{x}_{best} \leftarrow \mbox{arg}\min_{\vec{x}_t \in \mathcal{D}} \, y_t$
\STATE \textbf{return} the best uncovered input $\vec{x}_{best}$ and the corresponding function value $\mathcal{F}(\vec{x}_{best})$
\end{algorithmic}
\end{algorithm}

Consider the objective in \ref{eqn:main} written as a sum of two terms decomposed over the positive ($A^+$) and  non-positive ($A^-$) terms of matrix $A$:
\begin{align}
    \vec{x}^T A \vec{x} + b^T \vec{x} = \vec{x}^T A^+ \vec{x} + \vec{x}^T A^- \vec{x} + b^T \vec{x} \label{eqn:split}
\end{align}
where $A^+ + A^- = A$ and $A^+$ and $A^-$ are defined as follows:
\begin{align*}
  A^+ = \left\{
	\begin{array}{ll}
		a_{ij}  & \mbox{if } a_{ij} > 0 \\
		0 & \mbox{if } a_{ij} \leq 0 
	\end{array}
\right.  \forall a_{ij} \in A
\end{align*}
\begin{align*}
     A^- = \left\{
	\begin{array}{ll}
		a_{ij}  & \mbox{if } a_{ij} \leq 0 \\
		0 & \mbox{if } a_{ij} > 0 
	\end{array}
\right. \forall a_{ij} \in A
\end{align*}
The second term in Equation \ref{eqn:split} ($\vec{x}^T A^- \vec{x} + b^T \vec{x}$) is a submodular function. Similar to the strategy employed for prescriptive price optimization \cite{shinji_neurips}, we construct a submodular relaxation of the first term by bounding it below by a linear function $h(\vec{x})$ such that $h(\vec{x}) \leq \vec{x}^T A^+ \vec{x} \quad \forall \vec{x} \in \{0, 1\}^n$. It can be easily seen that $h(\vec{x}) = \vec{x}^T (A^+ \circ \Lambda) \vec{1} + \vec{1}^T (A^+ \circ \Lambda) \vec{x} - \vec{1}^T (A^+ \circ \Lambda) \vec{1}$ is an affine lower bound to $\vec{x}^T A^+ \vec{x}$, where $\circ$ represents Hadamard product and $\Lambda$ is a matrix defined as follows: $\Lambda = [\lambda_{ij}]_{n\times n}$, where $\lambda_{ij} \in [0, 1]$ is a parameter satisfying the following inequality:
\begin{align}
    \lambda_{ij} (x_i + x_j - 1) \leq x_i x_j 
\end{align}
 Using $h(\vec{x})$ as the affine lower bound, our new optimization problem becomes:
\begin{align}
    \min_{\vec{x} \in \mathcal{X}} h(\vec{x}) + \vec{x}^T A^- \vec{x} + b^T \vec{x} \label{relaxed_objective_separated}
\end{align}
We use $h_\Lambda(\vec{x})$ for denoting the combination of two linear terms in (\ref{relaxed_objective_separated}) i.e. $h_\Lambda(\vec{x}) = h(\vec{x}) + b^T \vec{x}$, along with signifying the dependence on $\Lambda$ parameter.
\begin{align}
    \min_{\vec{x} \in \mathcal{X}} h_\Lambda(\vec{x}) + \vec{x}^T A^- \vec{x} \label{relaxed_objective}
\end{align}
It should be noted again that the objective in (\ref{relaxed_objective}) is a lower bound of the original objective in (\ref{eqn:split}). This relaxed submodular objective can be solved exactly by turning it into a minimum graph cut problem and utilizing an efficient minimum graph cut algorithm \cite{boykov_kolmogorov_algorithm}. For a given ${\Lambda}$, we employ a standard graph construction strategy \cite{bqp_graph_cuts} dependent on the $\alpha$ parameters sampled from the surrogate model $f_{\alpha}$ at each BO iteration. A graph $G$ is constructed with $n+2$ vertices: $\mathcal{V} = \{s, t, v_1, \cdots, v_n\}$ where each non-terminal vertex $v_i$ encode one discrete variable $x_i \in \{0,1\}$. For each term depending on one variable $x_i$ (each non-zero entry of $h_\Lambda(\vec{x})$ in (\ref{relaxed_objective})), an edge is added in the graph from $s$ to $v_i$ with capacity $h_\Lambda(x_i)$ if $h_\Lambda(x_i)$ is positive or from $v_i$ to $t$ with capacity $-1 \cdot h_\Lambda(x_i)$ if it is negative. Further, each term depending on pair of variables $x_ix_j$ (each non-zero entry of $A^-$ in (\ref{relaxed_objective})) is represented by two edges i.e. edge $v_i$ to $v_j$  and edge $v_j$ to $t$ with capacity $-1\cdot A^-_{ij}$. 

\subsubsection{Optimizing Parameters to Improve Accuracy}

The quality of approximation of the above-mentioned submodular relaxation objective  (\ref{relaxed_objective}) critically depends on $\Lambda$ parameters. \cite{shinji_neurips} constructed an outer optimization problem to improve the accuracy of this relaxation and maximized the objective w.r.t $\Lambda$ to achieve the best approximation possible. We use the same procedure which is described below. By including the outer optimization over $\Lambda$, the overall problem becomes:
\begin{align}
    \max_{\Lambda \in [0, 1]^{n\times n}} \quad (\min_{\vec{x} \in D} h_\Lambda(\vec{x}) + \vec{x}^T A^- \vec{x} ) \label{outer_objective_original}
\end{align}
Equivalently, the outer maximization can be turned into minimization by considering the negative of the outer objective. 
\begin{align}
    \min_{\Lambda \in [0, 1]^{n\times n}} \quad -1\cdot (\min_{\vec{x} \in D} h_\Lambda(\vec{x}) + \vec{x}^T A^- \vec{x} ) \label{outer_objective}
\end{align}
This optimization problem can be solved efficiently using an iterative algorithm that alternates between solving the inner optimization over $\vec{x}$ via graph cut formulation and proximal gradient descent over $\Lambda$. If $\vec{x}_i$ is the solution of the submodularized inner objective in (\ref{outer_objective}) at the $i^{th}$ iteration for a fixed $\Lambda_i$, the update equation for $\Lambda$ is given as follows:
\begin{align}
    \Lambda_{i+1} = (\Lambda_i - \eta_i G_i)^{\perp}
\end{align}
where $\eta_i$ is the step size, $G_i$ is the sub-gradient of the outer objective in (\ref{outer_objective}) defined as $G_i = A^+ \circ (\vec{1}\vec{1}^T - \vec{x}_i \vec{1}^T - \vec{1}\vec{x}_i^T)$  and $\perp$ is the projection operator for any matrix $P$ defined as $P_{ij}^{\perp} = \{0 \text{ if } P_{ij} < 0, 1 \text{ if } P_{ij} > 1$, and $P_{ij}$ otherwise\} . We employ proximal gradient descent because it scales gracefully, is amenable to recent advances in auto-differentiation tools, and fast convergence \cite{proximal_gradient_descent}.
We require few iterations (~5-10) of proximal gradient descent and each inner optimization is very fast because of strongly polynomial graph cut algorithms. Indeed, our experiments validate this claim over multiple real-world benchmarks.

%% file: files/experiments.tex
\section{Experiments and Results}
In this section, we describe our experimental setup, and present results comparing state-of-the-art BOCS method with SDP relaxation and our proposed parameterized submodular relaxation (PSR) algorithm.

\subsection{Experimental Setup}
{\noindent \bf Benchmark domains.} We employ four diverse synthetic and real-world benchmarks for our empirical evaluation.

\vspace{1.0ex}

{\bf 1. Binary quadratic programming (BQP)}. The goal in binary quadratic programming (BQP) \cite{bocs} is to maximize a binary quadratic function with $l_1$ regularization: $\max_{\vec{x} \in \{0,1\}^n} (\vec{x}^T Q \vec{x} - \lambda \|\vec{x}\|_1)$, where $Q $ is a randomly generated matrix defined as Hadamard product of two matrices ($M$ and $K$);  $Q = M \circ K$, where $M \in \mathcal{R}^{n \times n}, M_{ij} = \mathcal{N}(0, 1)$, $\mathcal{N}(0, 1)$ stands for the standard Gaussian distribution and $K \in \mathcal{R}^{n \times n}, K_{ij} = \exp(-(i-j)^2/\alpha^2)$, $\alpha$ is the correlation length parameter. 

\vspace{1.0ex}

{\bf 2. Contamination.} This problem considers a food supply with $n$ stages, where a binary \{0,1\} decision ($x_i$) must be made at each stage to prevent the food from being contaminated with pathogenic micro-organisms \cite{contamination}. Each prevention effort at stage $i$ can be made to decrease the contamination by a given random rate $\Gamma_i$ and incurring a cost $c_i$. The contamination spreads with a random rate $\Lambda_i$ if no prevention effort is taken. The overall goal is to ensure that the fraction of contaminated food at each stage $i$ does not exceed an upper limit $U_i$ with probability at least $1-\epsilon$ while minimizing the total cost of all prevention efforts. Following \cite{bocs}, the lagrangian relaxation based problem formulation is given below:
\begin{align*}
    \arg \min_x \sum_{i=1}^n \left[c_i x_i + \frac{\rho}{T} \sum_{k=1}^T 1_{\{Z_k > U_i\}} \right] + \lambda \|x\|_1
\end{align*}
where $\lambda$ is a regularization coefficient, $Z_i$ is the fraction of contaminated food at stage $i$, violation penalty coefficient $\rho$=$1$, and $T$=$100$.

\vspace{1.0ex}

{\bf 3. Sparsification of zero-field Ising models (Ising).} The distribution of a  zero field Ising model $p(z)$ for $z \in \{-1, 1\}^n$ is characterized by a symmetric interaction matrix $J^p$ whose support is represented by a graph $G^p = ([n], E^p)$ that satisfies $(i,j) \in E^p$ if and only if $J^p_{ij} \neq 0$ holds \cite{bocs}. The overall goal in this problem is to find a close approximate distribution $q(z)$ while minimizing the number of edges in $E^q$. Therefore, the objective function in this case is a regularized KL-divergence between $p$ and $q$ as given below:
\begin{align*}
    D_{KL} (p || q_{\vec{x}}) = \sum_{(i,j) \in E^p} (J_{ij}^p - J_{ij}^q) E_p[z_iz_j] + log (Z_q/Z_p)
\end{align*}
where $Z_q$ and $Z_p$ are partition functions corresponding to $p$ and $q$ respectively, and $\vec{x} \in \{0,1\}^{E^q}$ is the decision variable representing whether each edge is present in $E^q$ or not.

\vspace{1.0ex}

{\bf 4. Low auto-correlation binary sequences (LABS).} The problem is to find a binary \{+1,-1\} sequence $S$ = $(s_1, s_2,\cdots, s_n)$ of given length $n$ that maximizes {\it merit factor} defined over a binary sequence as given below:
\begin{align*}
\text{Merit Factor(S)} &= \frac{n^2}{E(S)}
\hspace{1mm}\\
\text{where} \hspace{1mm} E(S) =& \sum_{k=1}^{n-1} \left(\sum_{i=1}^{n-k} s_i s_{i+k}\right)^2
\end{align*}
The LABS problem has multiple applications in diverse scientific disciplines \cite{LABS}.

\vspace{1.0ex}

\noindent {\bf Algorithmic setup.} For the sake of consistency, we convert all benchmark problems to minimization. Note that this can be achieved by minimizing the negative of the original objective if the true goal is to maximize the objective. We built our code on top of the open-source Python implementation of BOCS \footnote{\url{https://github.com/baptistar/BOCS}}. We employed the Boykov-Kolmogorov algorithm from graph-tool library \footnote{\url{https://graph-tool.skewed.de/}} for solving minimum graph cut formulation of the relaxed submodular acquisition function objective noting that any minimum cut algorithm can be used to the same effect. We employed two initializations (random and $\vec{1}\vec{1}^T/2$) for optimizing $\Lambda$ parameter within PSR noting that both gave similar results. We ran proximal gradient descent procedure for a maximum of 10 iterations on all benchmarks and achieved convergence.
All the reported results are averaged over 10 random runs.

\vspace{1.5ex}

{\noindent \bf Evaluation metrics.} We demonstrate the advantages of our PSR algorithm for combinatorial BO by comparing it with the state-of-the-art BOCS approach along two fronts. 

\vspace{1.0ex}

{\bf 1) Scalability and accuracy of AF optimization.} We compare PSR and SDP approaches for solving acquisition function optimization problems within BOCS method. To evaluate  scalability for AFO, we report the average AFO time across all BO iterations normalized w.r.t PSR. Suppose $T_{SDP}$ and $T_{PSR}$ stand for average AFO time for some input dimensionality $d$. We normalize $T_{SDP}$ and $T_{PSR}$ using the AFO time of PSR for the smallest dimension. To evaluate the accuracy of solving AFO problems, we report the average percentage improvement in the AF objective achieved by PSR when compared to the corresponding AF objective from SDP. 

\begin{figure*}[t!]
\centering
\subfloat[Subfigure 2 list of figures text][BQP]{
\includegraphics[width=0.5\textwidth]{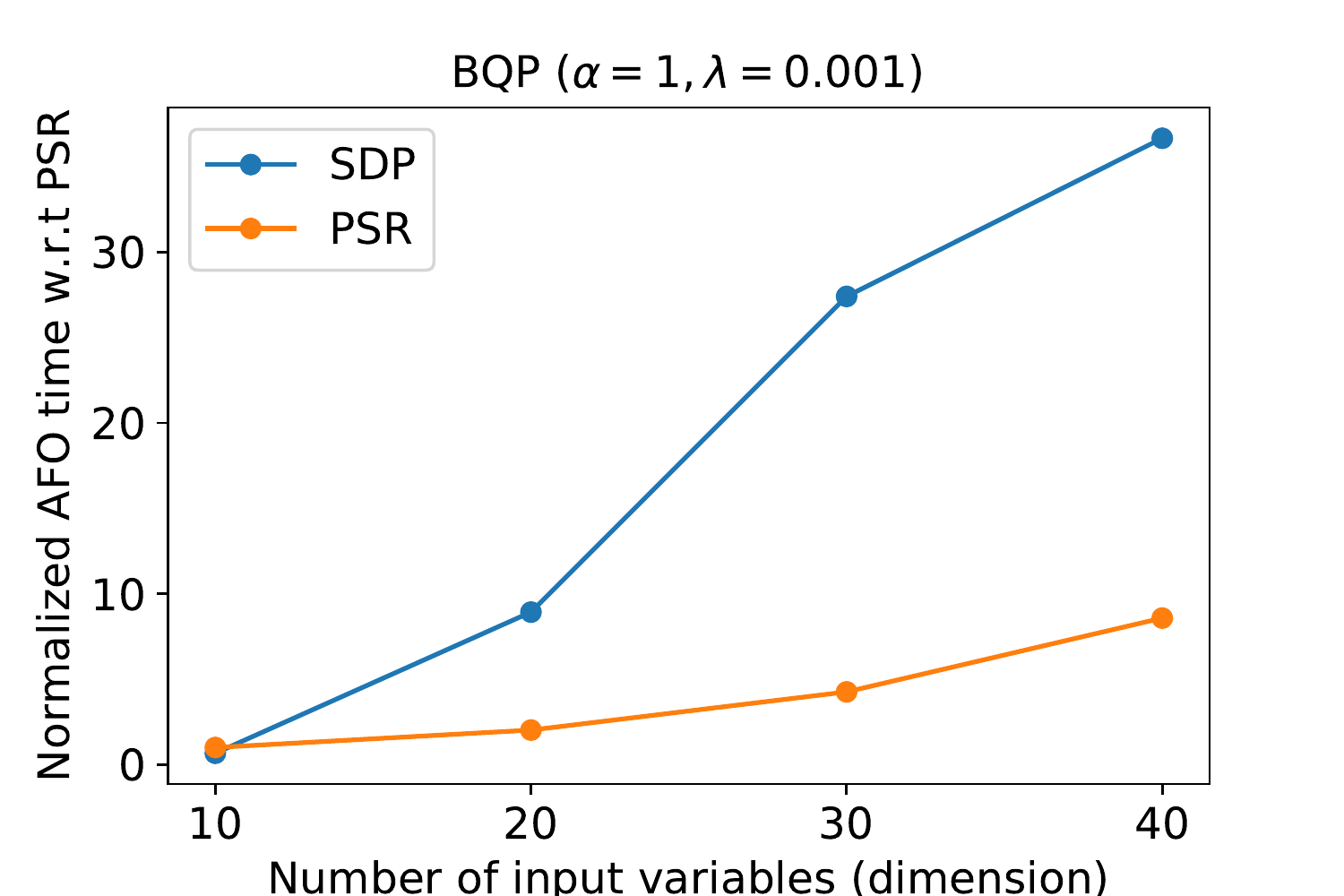}
\label{fig:at_bqp}}
\subfloat[Subfigure 1 list of figures text][Contamination]{
\includegraphics[width=0.5\textwidth]{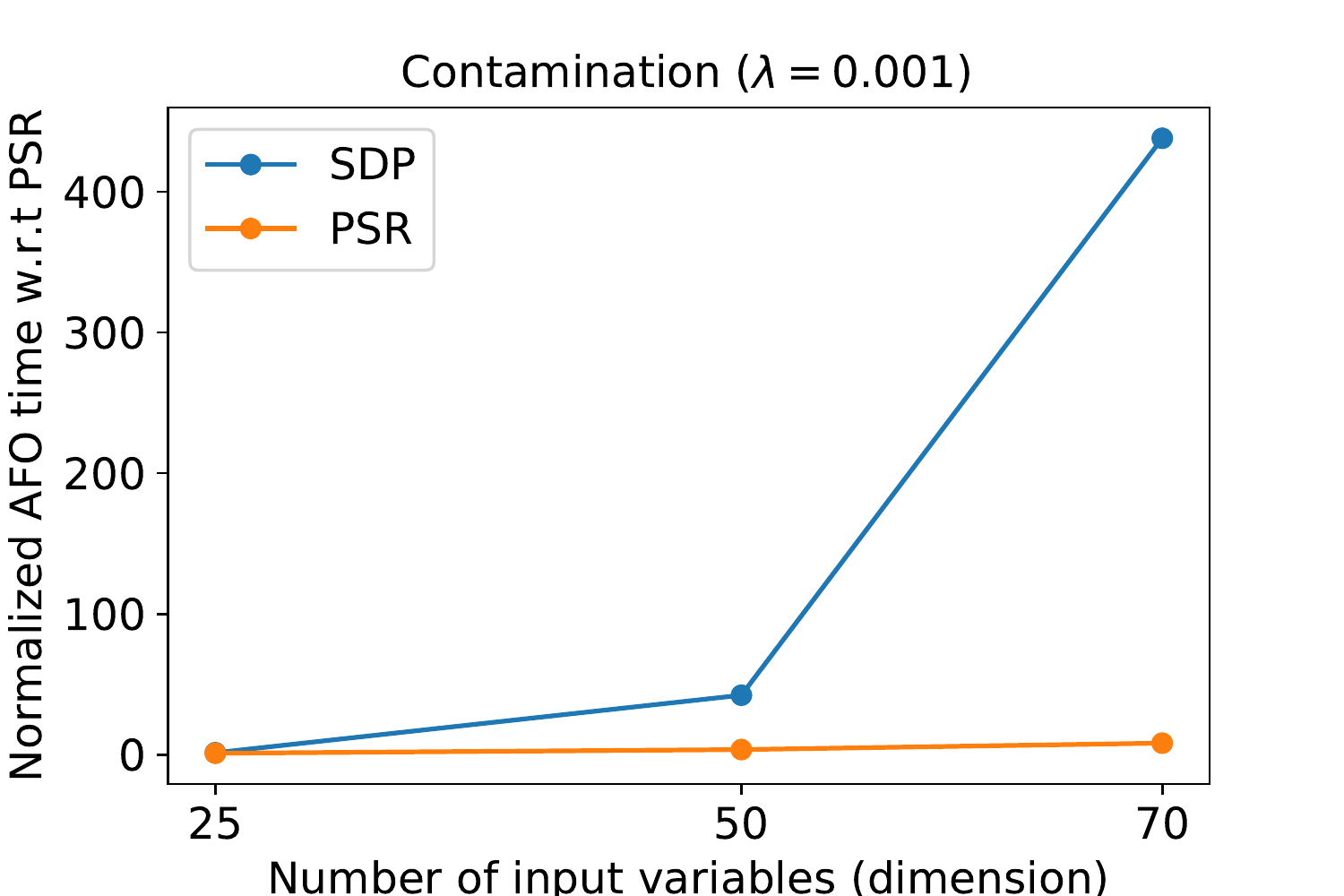}
\label{fig:at_cont}}
\vspace{-3ex}
\quad
\subfloat[Subfigure 2 list of figures text][Ising]{
\includegraphics[width=0.5\textwidth]{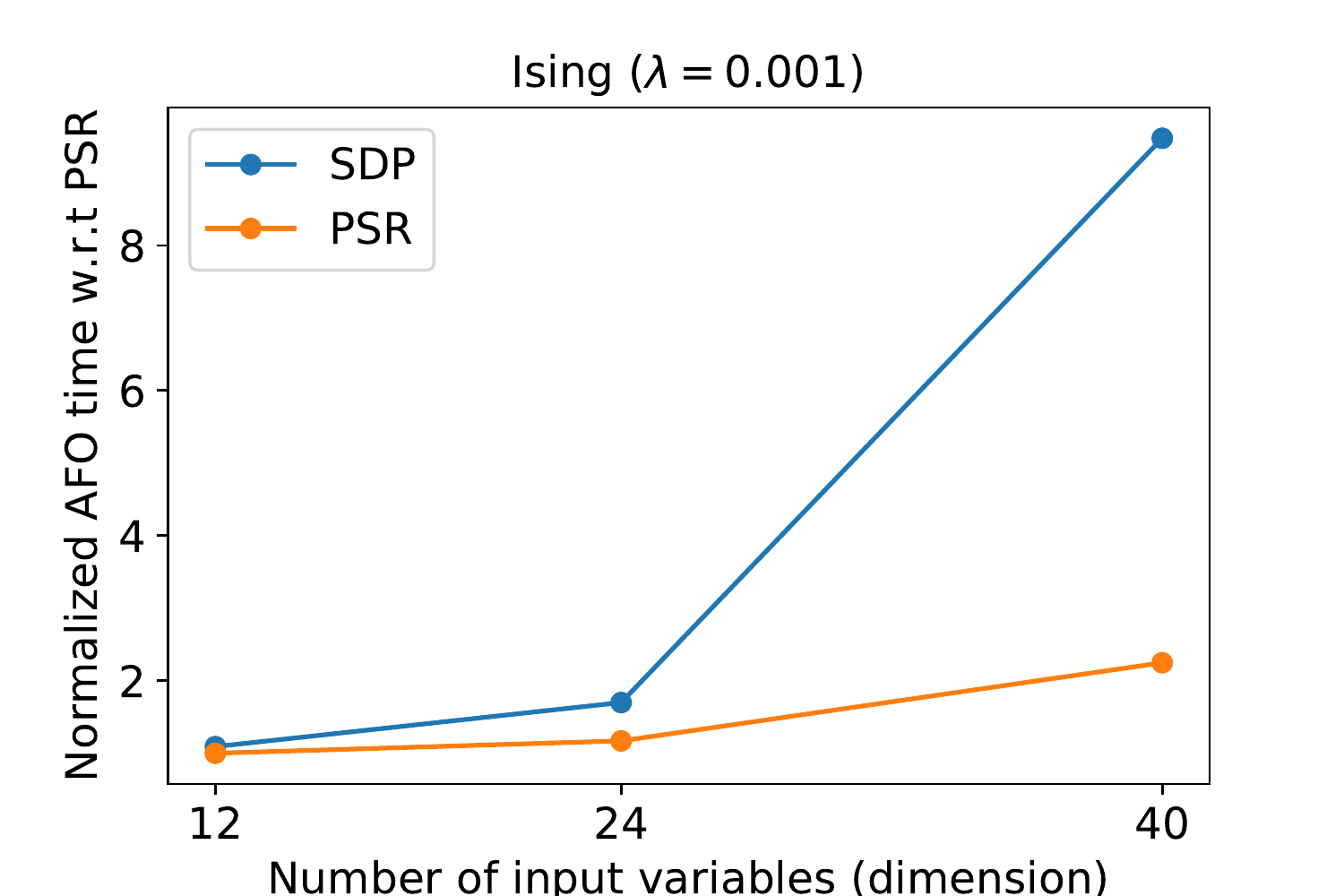}
\label{fig:at_ising}}
\subfloat[Subfigure 1 list of figures text][LABS]{
\includegraphics[width=0.5\textwidth]{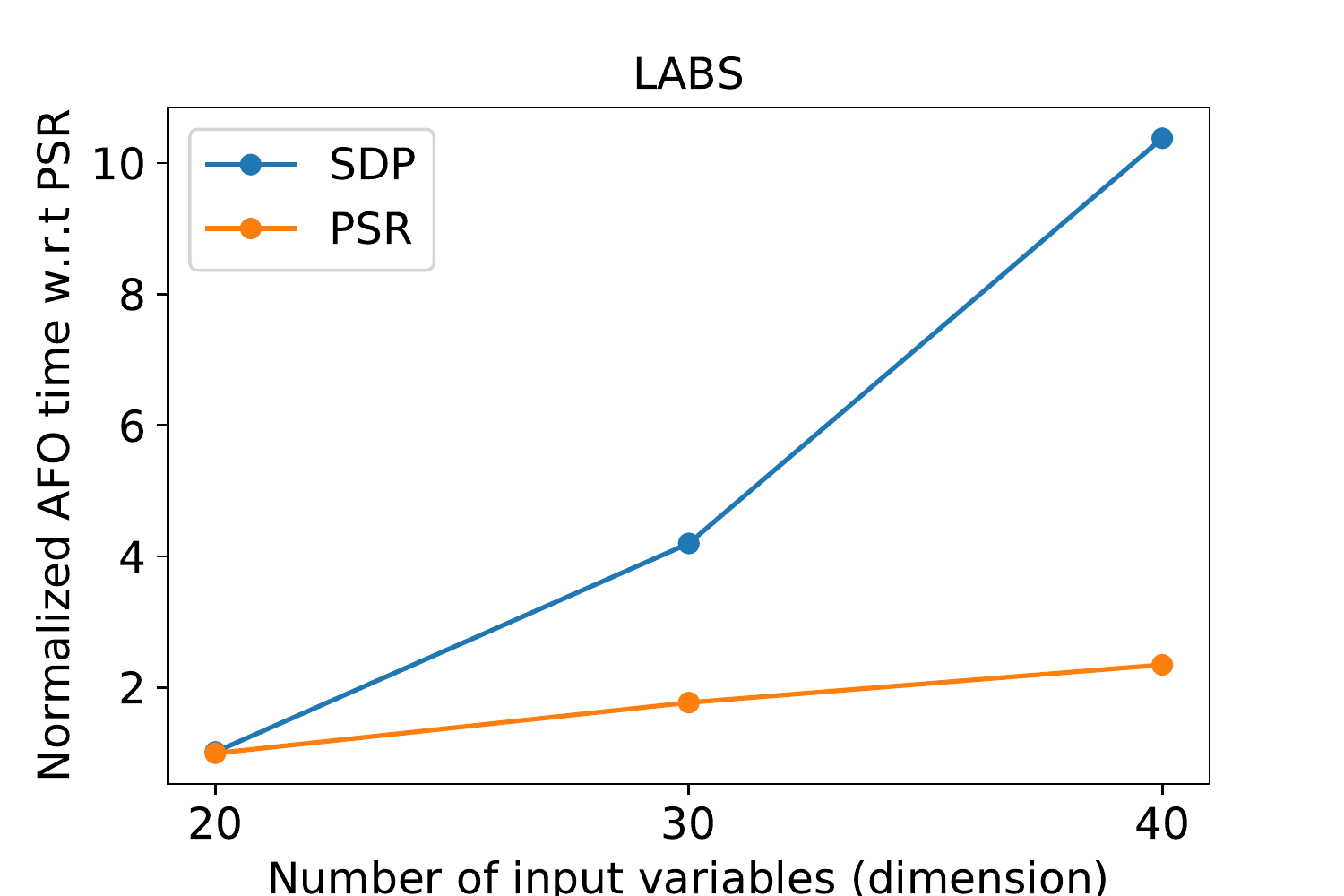}
\label{fig:at_labs}}
\caption{Results comparing PSR algorithm and SDP approach on {\em average AFO time} normalized w.r.t PSR. The title of each figure refers to the benchmark with corresponding parameter (if any).} 
\label{fig:at}
\end{figure*}

\begin{figure*}[t]
\centering
\subfloat[Subfigure 2 list of figures text][BQP]{
\includegraphics[width=0.5\textwidth]{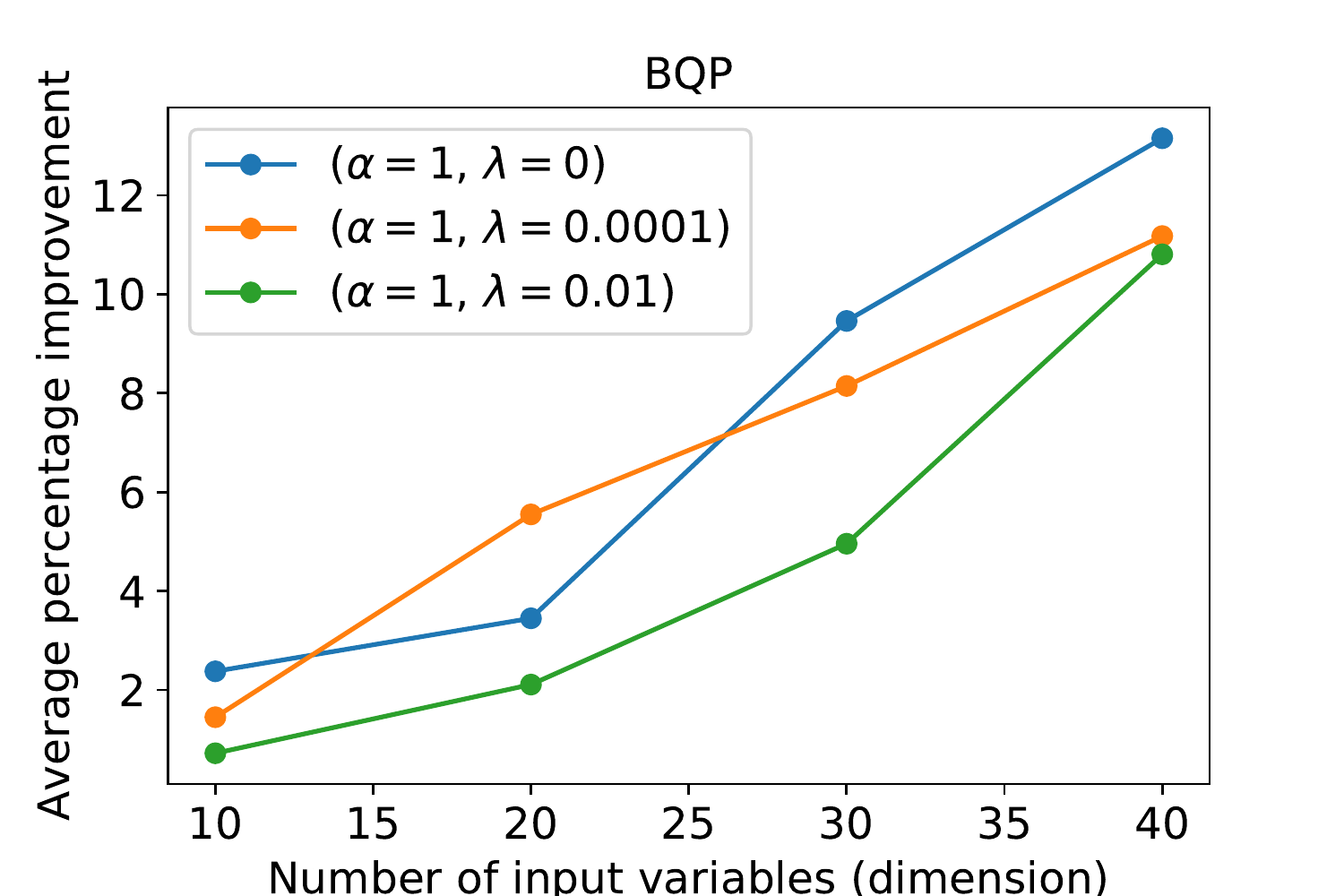}
\label{fig:ap_bqp}}
\subfloat[Subfigure 1 list of figures text][Contamination]{
\includegraphics[width=0.5\textwidth]{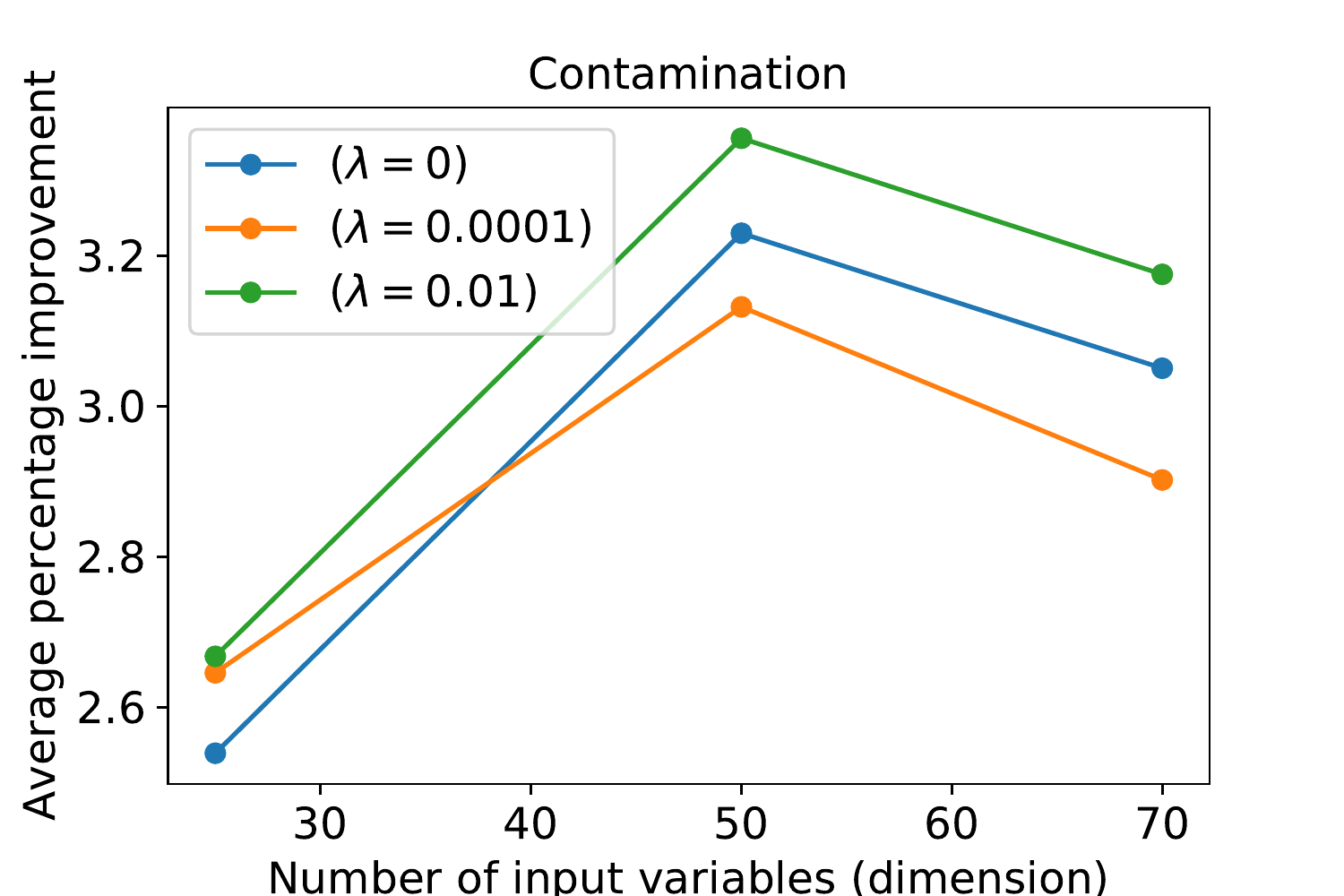}
\label{fig:ap_cont}}
\vspace{-3ex}
\quad
\subfloat[Subfigure 2 list of figures text][Ising]{
\includegraphics[width=0.5\textwidth]{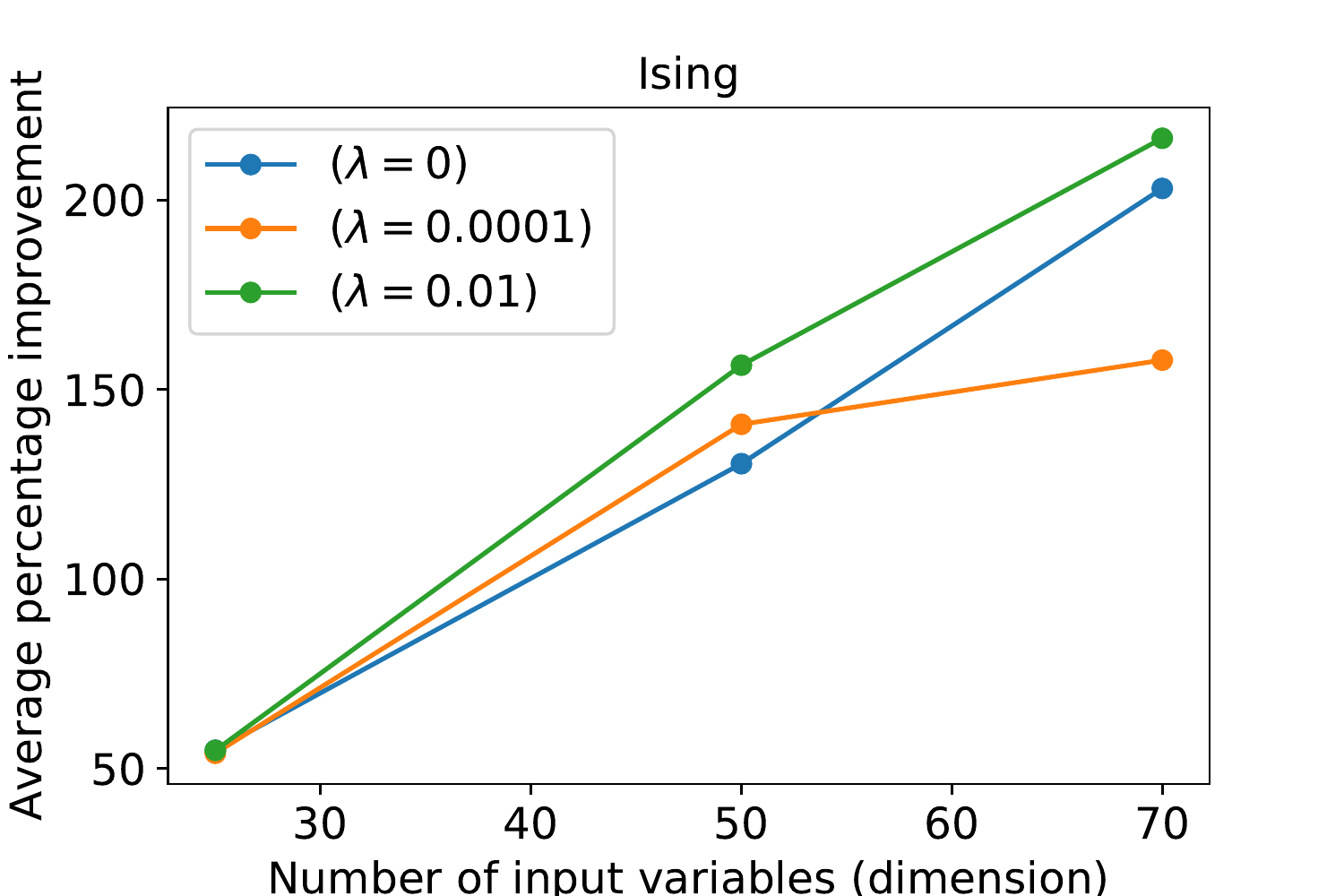}
\label{fig:ap_ising}}
\subfloat[Subfigure 1 list of figures text][LABS]{
\includegraphics[width=0.5\textwidth]{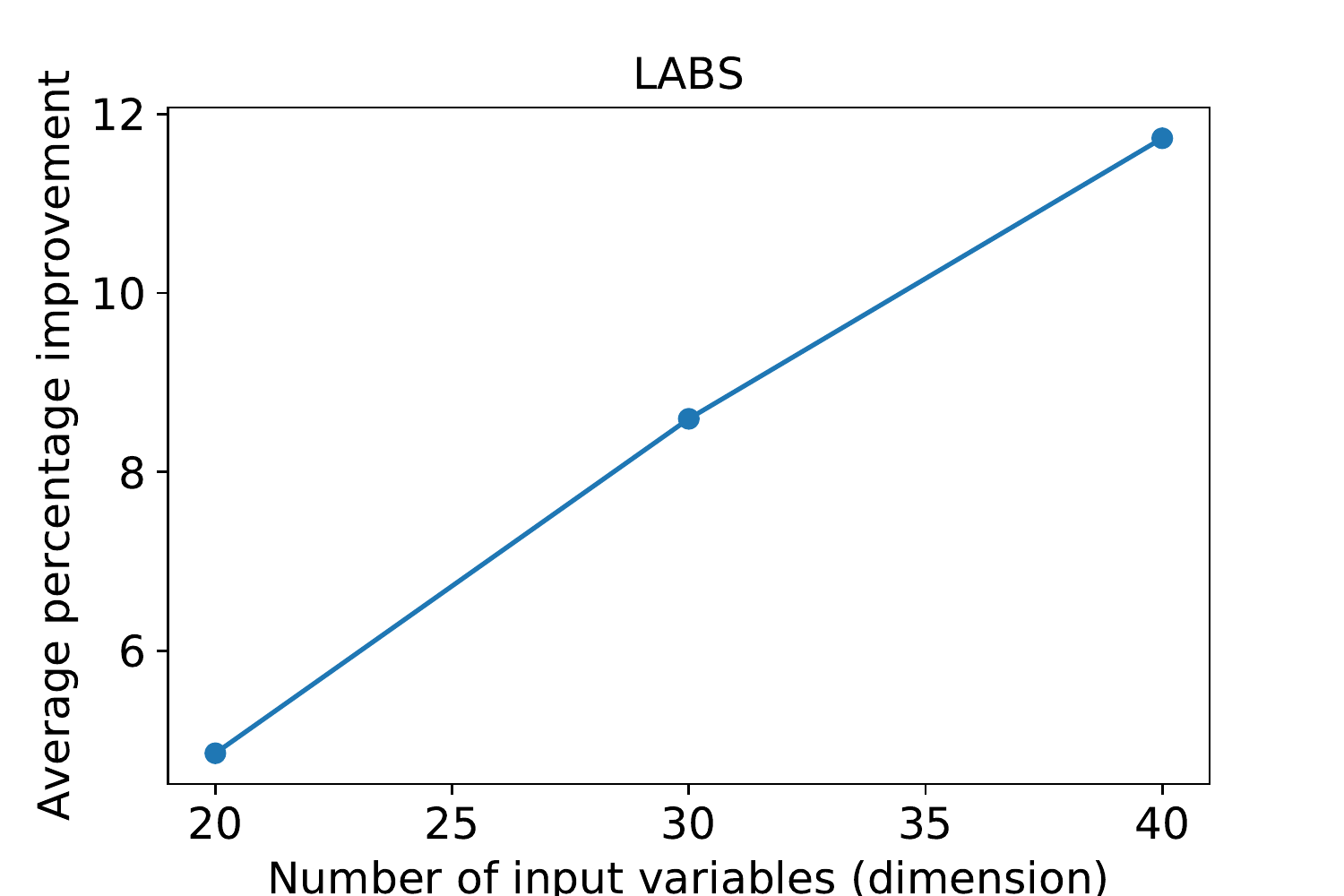}
\label{fig:ap_labs}}
\caption{Results comparing the accuracy of the AF minimizers obtained by PSR and SDP approach measured in terms of {\em Average percent improvement} in the AF objective over all iterations of the BO procedure.} 
\vspace{-2.5ex}
\label{fig:ap}
\end{figure*}
\begin{figure*}[h!]
\centering
\subfloat[Subfigure 2 list of figures text][BQP]{
\includegraphics[width=0.5\textwidth]{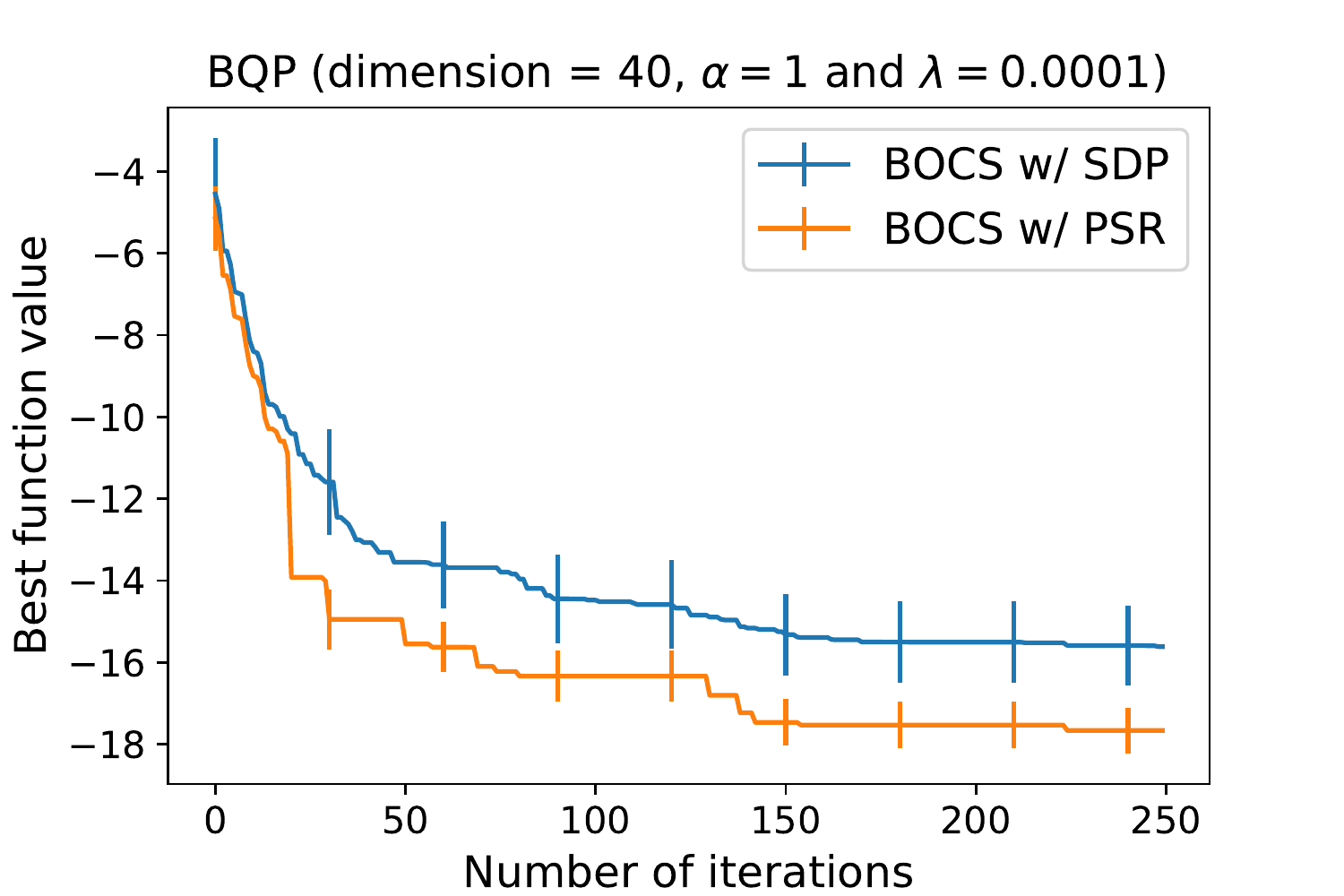}
\label{fig:bo_bqp}}
\subfloat[Subfigure 1 list of figures text][Contamination]{
\includegraphics[width=0.5\textwidth]{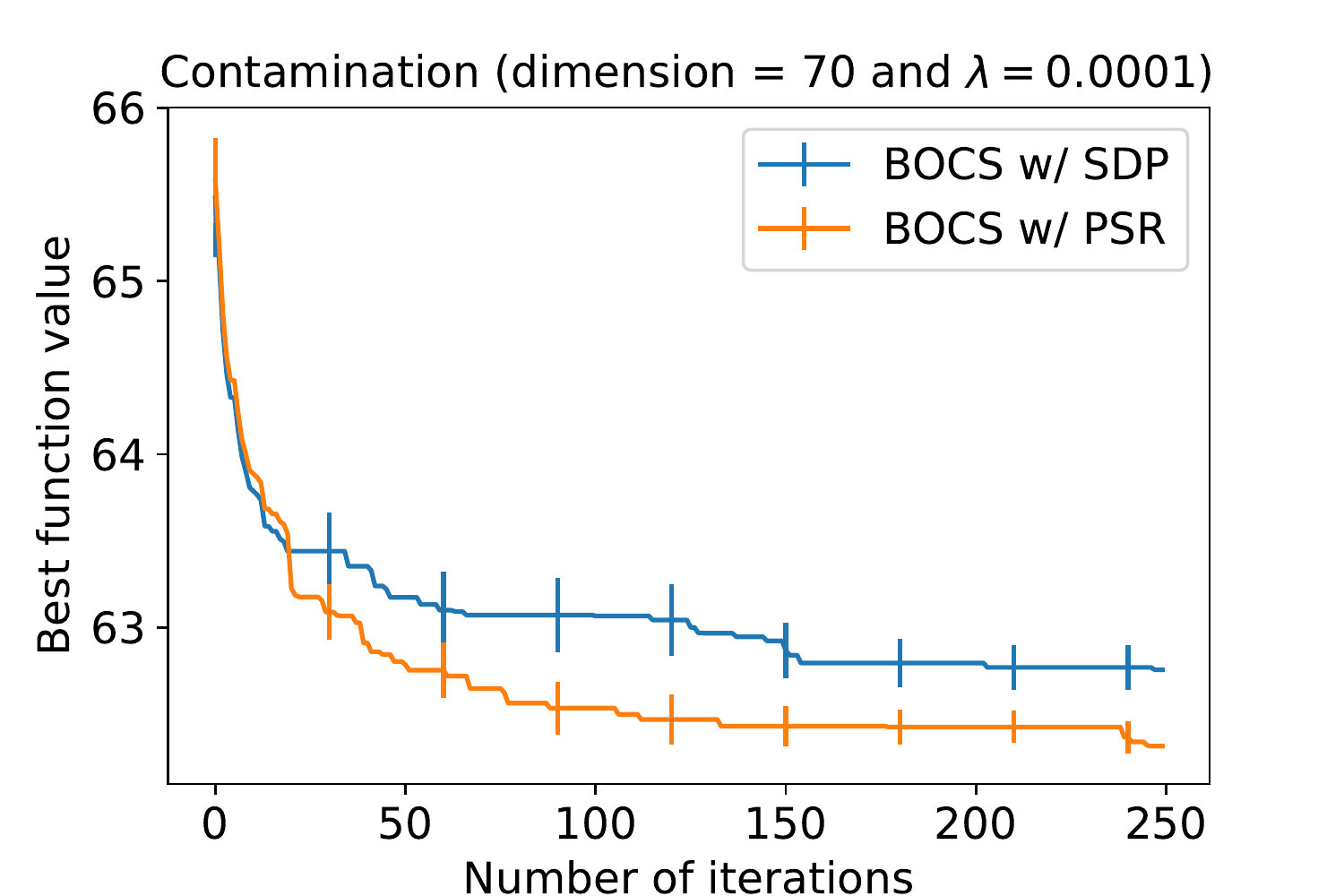}
\label{fig:bo_cont}}
\quad
\subfloat[Subfigure 2 list of figures text][Ising]{
\includegraphics[width=0.5\textwidth]{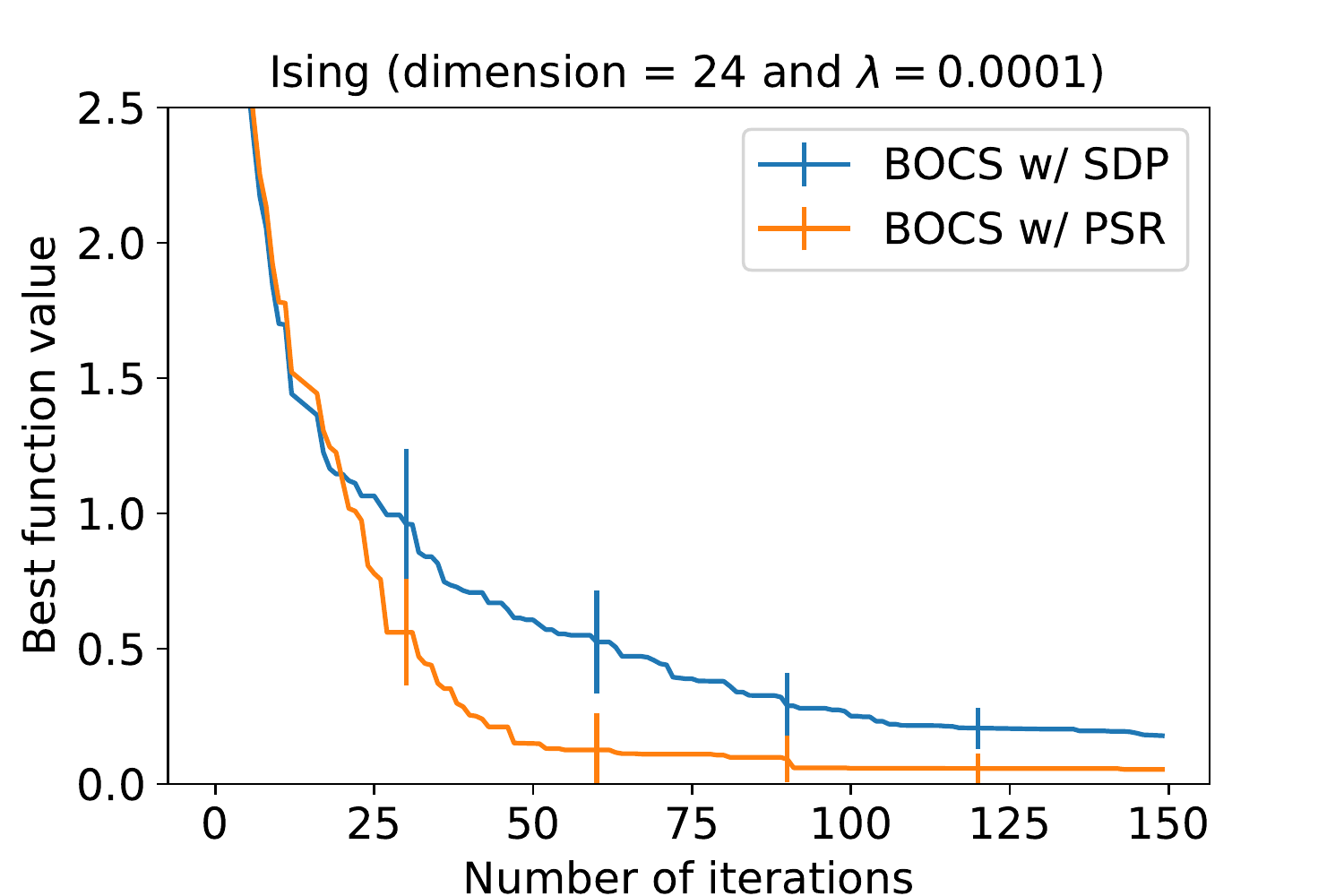}
\label{fig:bo_ising}}
\subfloat[Subfigure 1 list of figures text][LABS]{
\includegraphics[width=0.5\textwidth]{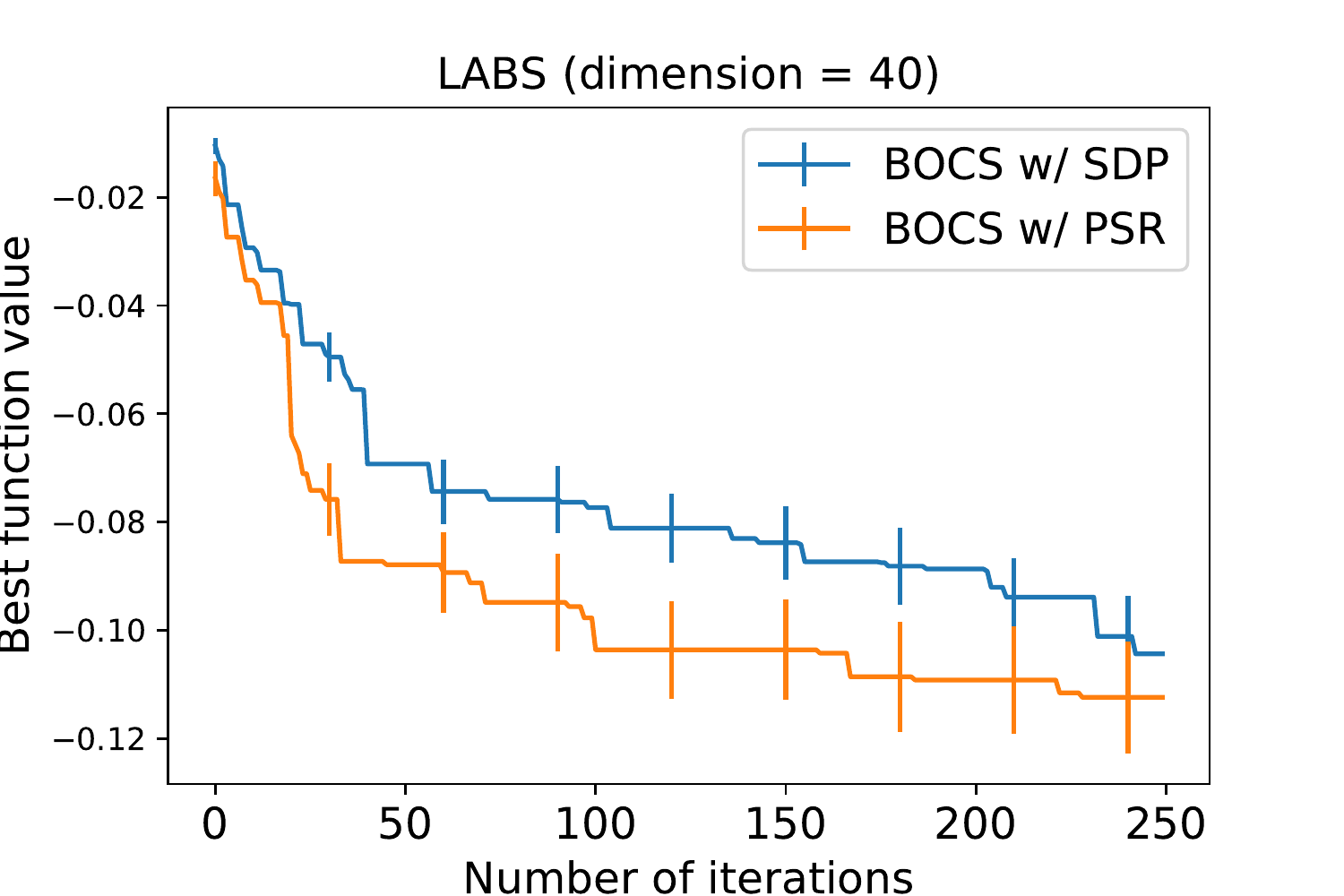}
\label{fig:bo_labs}}
\caption{Results comparing BOCS with PSR algorithm and BOCS with SDP approach on {\em best function value achieved} versus number of iterations. The horizontal axis (x-axis) depicts number of iterations while the vertical axis (y-axis) depicts the {\em best function value}.  } 
\label{fig:bo}
\end{figure*}

\vspace{1.0ex}

{\bf 2) Overall BO accuracy.} We use the best function value achieved after a given number of BO iterations (function evaluations) as a metric to evaluate the two methods: \texttt{BOCS w/ SDP} and \texttt{BOCS w/ PSR}. Note that BOCS is already shown to significantly improve over SMAC \cite{bocs}. The method that uncovers high-performing combinatorial structures with less number of function evaluations is considered better.
We use the total number of BO iterations similar to BOCS \cite{bocs}.

\subsection{Results for Acquisition Function Optimization}

We present a canonical result for each benchmark domain in Figure \ref{fig:at} and \ref{fig:bo} by fixing the objective parameter (if any) to a particular value, e.g., $\alpha = 1$, $\lambda = 0.001$ for BQP and $\lambda = 0.0001$ for Contamination and Ising in Figure \ref{fig:at} while noting that {\em all} experiments followed the same pattern.

\vspace{1.0ex}

{\noindent \bf Average AFO time.} Figure \ref{fig:at} shows the results of PSR and SDP approaches as a function of increasing dimension. Recall that we normalize the average AFO time w.r.t that of PSR for smallest dimension (base case). We can clearly see that the proposed PSR approach requires significantly low computation time when compared to the SDP approach and the gap increases with increasing input dimensions. This supports our claim that PSR algorithm improves the scalability of AFO problems in combinatorial BO setting.  It should be noted that AFO problem is solved at each BO iteration. For example, if we run BO for 250 iterations, we need to solve 250 AFO problems. Therefore, the computational-efficiency of PSR is compounded across the entire BO procedure. 

\vspace{1.0ex}

{\noindent \bf Average percentage improvement in AF objective.} PSR algorithm also finds better optimized value for AFO problems on each benchmark domain as shown in Figure \ref{fig:ap}. The vertical axis of the plots in Figure \ref{fig:ap} represent the average percentage improvement in AF objective obtained by PSR when compared to that obtained by SDP (higher the better). PSR algorithm always finds a minimizer with lower AF value when compared to SDP's minimizer on all benchmarks. Furthermore, this accuracy gap increases with increasing dimensions reinforcing the ability of PSR to scale to large dimensions while also improving the accuracy.

\subsection{Results for Overall BO Accuracy}

The main goal in BO is to find best accuracy on the true expensive black-box function $\mathcal{O}$. Ideally, the gains in accuracy for solving AFO problems as shown in previous section should reflect in the overall BO performance using the proposed PSR approach. Indeed, Figure \ref{fig:bo} clearly shows that using the BOCS model with PSR algorithm improves the overall accuracy of the BO procedure on all benchmark domains. This is a direct consequence of the improved accuracy achieved by the PSR algorithm in solving AFO problems at each BO iteration. All the reported results are averaged over 10 random runs.